\documentclass[runningheads]{llncs}
\usepackage{graphicx}
\usepackage{amsmath,amssymb} 
\usepackage{color}
\usepackage{blindtext}
\usepackage{multirow}
\usepackage[width=122mm,left=12mm,paperwidth=146mm,height=193mm,top=12mm,paperheight=217mm]{geometry}
\usepackage{authblk}

\begin{document}
\pagestyle{headings}
\mainmatter

\title{Modular Generative Adversarial Networks} 
\titlerunning{Modular Generative Adversarial Networks}
\authorrunning{Bo Zhao, Bo Chang, Zequn Jie and Leonid Sigal}

\author{Bo Zhao$^*$\qquad Bo Chang$^*$\qquad Zequn Jie$^\dagger$\qquad Leonid Sigal$^*$}


\institute{$^*$University of British Columbia\qquad $^\dagger$Tencent AI Lab\\
	\email{\{bzhao03, lsigal\}@cs.ubc.ca\qquad bchang@stat.ubc.ca\qquad zequn.nus@gmail.com}
}

\maketitle

\begin{figure}[!h]
\begin{center}
\includegraphics[width=1\columnwidth]{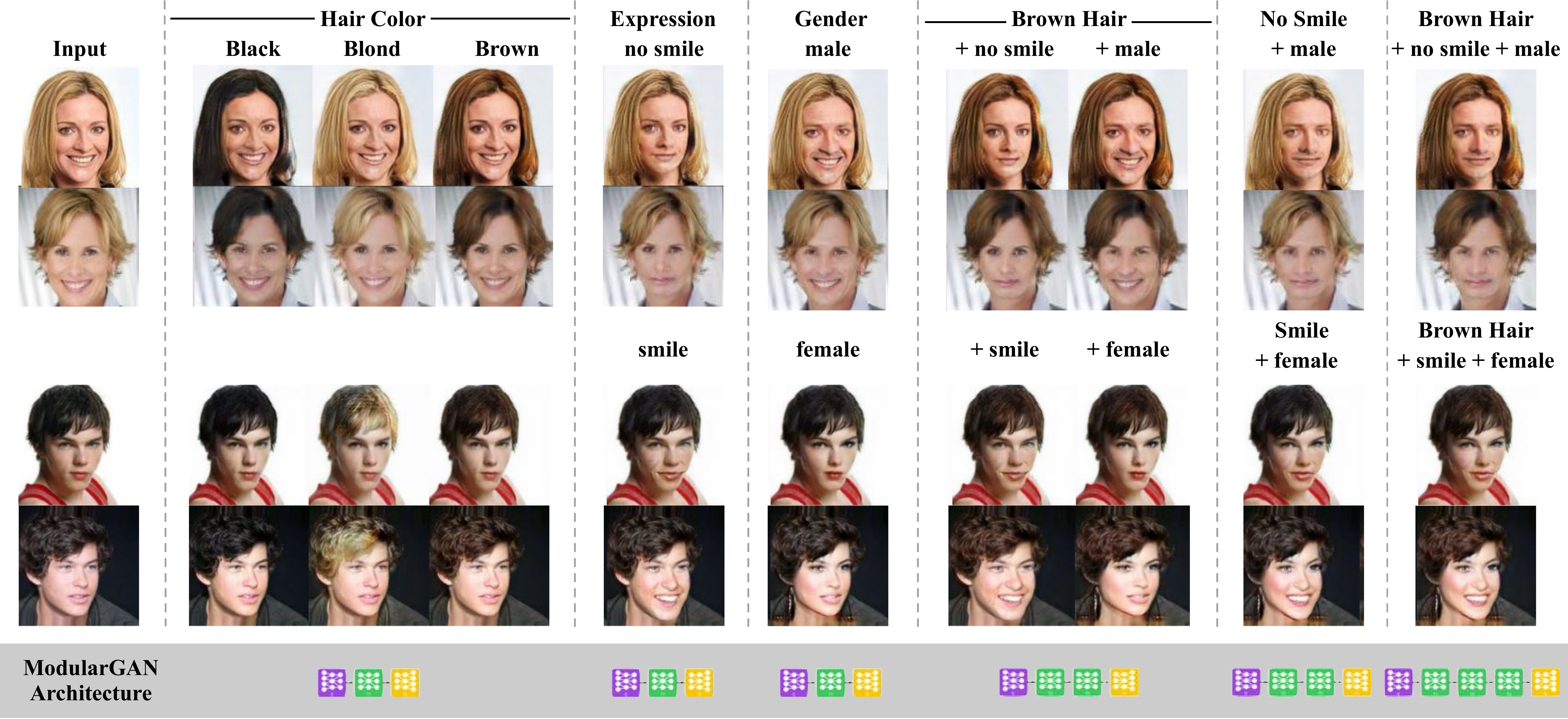}
\end{center}
\vspace{-0.2in}
\caption{{\bf ModularGAN:} Results of proposed modular generative adversarial network illustrated on multi-domain image-to-image translation task on the CelebA~\cite{Liu2015} dataset.}
\vspace{-0.2in}
\label{fig:results_ours}
\end{figure}
\vspace{-0.2in}

\begin{abstract}
Existing methods for multi-domain image-to-image translation (or generation) attempt to directly map an input image (or a random vector) to an image in one of the output domains. However, most existing methods have limited scalability and robustness, since they require building independent models for each pair of domains in question. This leads to two significant shortcomings: (1) the need to train exponential number of pairwise models, and (2) the inability to leverage data from other domains when training a particular pairwise mapping. 
Inspired by recent work on module networks \cite{Andreas2016a}, this paper proposes ModularGAN for multi-domain image generation and image-to-image translation. 
ModularGAN consists of several reusable and composable modules that carry on different functions (e.g., encoding, decoding, transformations). 
These modules can be trained simultaneously, leveraging data from all domains, and then combined to construct specific GAN networks at test time, according to the specific image translation task. 
This leads to ModularGAN's superior flexibility of generating (or translating to) an image in any desired domain.
Experimental results demonstrate that our model not only presents compelling perceptual results but also outperforms state-of-the-art methods on multi-domain facial attribute transfer.

\end{abstract}

\section{Introduction}
Image generation has gained popularity in recent years following the introduction of variational autoencoder~\cite{Kingma2013} and generative adversarial networks~\cite{Goodfellow2014}. A plethora of tasks, based on image generation, have been studied, including attribute-to-image generation~\cite{Mirza2014,Odena2016,Yan2016}, text-to-image generation~\cite{Reed2016,Reed2016a,Xu2017,Zhang2017,Zhao2017} or image-to-image translation~\cite{Isola2016,Karacan2016,Sangkloy2016,Zhu2017,Choi2017}. These tasks can be broadly termed {\em conditional image generation}, which takes an attribute vector, text description or an image as the conditional input, respectively, and outputs an image.  Most existing conditional image generation models learn a direct mapping from inputs, which can include an image or a random noise vector, and target condition to output an image containing target properties, using a neural network. 

Each condition, or condition type, effectively defines a generation or image-to-image output domain ({\em e.g.}, domain of {\em expression} (smiling) or {\em gender} (male / female) for facial images). For practical tasks, it is desirable to be able to control a large and variable number of conditions ({\em e.g.}, to generate images of person {\em smiling} or {\em brown haired smiling man}). Building a function that can deal with the exponential, in the number of conditions, domains is difficult. Most existing image translation methods \cite{Isola2016,Karacan2016,Sangkloy2016,Zhu2017} can only translate images from one domain to another. For multi-domain setting this results in a number of shortcomings: (i) requirement to learn an exponential number of pairwise translation functions, which is computationally expensive and practically infeasible for more than a handful of conditions; (ii) it is impossible to leverage data from other domains when learning a particular pairwise mapping; and (iii) the pairwise translation function could potentially be arbitrarily complex in order to model the transformation between very different domains. To address (i) and (ii), multi-domain image (and language \cite{Johnson2017a}) translation \cite{Choi2017} models have been introduced very recently. A fixed vector representing the source/target domain information can be used as the condition for a single model to guide the translation process. However, the sharing of information among the domains is largely implicit and the functional mapping becomes even more excessively complex. 

We posit that dividing the image generation process into multiple simpler generative steps can make the model easier and more robust to learn. In particular, we neither train pairwise mappings \cite{Isola2016,Zhu2017} nor one complex model \cite{Perarnau2016,Choi2017}; instead we train a small number of simple generative modules that can compose to form complex generative processes. In particular, consider transforming an image from domain $A$ ({\em man frowning}) to $C$ ({\em woman smiling}): $\mathcal{D}_A \rightarrow \mathcal{D}_C$. It is conceivable, even likely, that first transforming the original image to depict a {\em female} and subsequently {\em smiling} ($\mathcal{D}_A \xrightarrow{female} \mathcal{D}_B \xrightarrow{smiling} \mathcal{D}_C$) would be more robust than directly going from domain $A$ to $C$. The reason is two fold: (i) the individual transformations are simpler and spatially more local, and (ii) the amount of data in the intermediate {\em female} and {\em smile} domains are by definition larger than in the final domain of {\em woman smiling}. In other words, in this case, we are leveraging more data to learn simpler translation/transformation functions. This intuition is also consistent with recently introduced modular networks \cite{Andreas2016a,Andreas2016}, which we here conceptually adopt and extend for generative image tasks.

To achieve and formalize  this incremental image generation process, we propose the modular generative adversarial network (ModularGAN). 
ModularGAN consists of several different modules, including {\em generator}, {\em encoder}, {\em reconstructor}, {\em transformer} and {\em discriminator}, trained jointly. 
Each module performs specific functionality.
The {\em generator} module, used in image generation tasks, generates a latent representation of the image from a random noise and an (optional) condition vector.
The {\em encoder} module, used for image-to-image translation, encodes the input image into a latent representation.
The latent representation, produced by either generator or encoder, is manipulated by the the {\em transformer} module according to the provided condition.
The {\em reconstructor} module then reconstructs the transformed latent representation to an image.
The {\em discriminator} module is used to distinguish whether the generated or transformed image looks real or fake, and also to classify the attributes of the image.
Importantly, different {\em transformer} modules can be composed dynamically at test time, in any order, to form generative networks that apply a sequence of feature transformations in order to obtain more complex mappings and generative processes. 

\vspace{0.1in}
\noindent
{\bf Contributions:}
Our contributions are multi-fold, 
\vspace{-0.1in}
\begin{enumerate}
  \item[-] We propose ModularGAN -- a novel modular multi-domain generative adversarial network architecture. 
  ModularGAN consists of several reusable and composable modules. Different modules can be combined easily at test time, in order to generate/translate an image in/to different domains efficiently. To the best of our knowledge, this is the {\em first} modular GAN architecture.
  \item[-] We provide an efficient way to train all the modules jointly end-to-end. New modules can be easily added to our proposed ModularGAN, and a subset of the existing modules can also be upgraded without affecting the others.
  \item[-] We demonstrate how one can successfully combine different (transformer) modules in order to translate an image to different domains. We utilize mask prediction, in the transformer module, to ensure that only local regions of the feature map are transformed; leaving other regions unchanged.
  \item[-] We empirically demonstrate the effectiveness of our approach on image generation (ColorMNIST dataset) and image-to-image translation (facial attribute transfer) tasks. Qualitative and quantitative comparisons with state-of-the-art GAN models illustrate improvements obtained by ModularGAN.
\end{enumerate}

\vspace{-0.25in}
\section{Related work}
\vspace{-0.1in}
\subsection{Modular Networks}
\vspace{-0.1in}
Visual question answering (VQA) is a fundamentally compositional task. 
By explicitly modeling its underling reasoning process, \textbf{Neural module networks}~\cite{Andreas2016a} are constructed to perform various operations, including attention, re-attention, combination, classification, and measurement.
Those modules are assembled into all configurations necessary for different question tasks.
A natural language parser decompose questions into logical expressions and dynamically lay out a deep network composed of reusable modules.
\textbf{Dynamic neural module networks}~\cite{Andreas2016} extend neural module networks by learning the network structure via reinforcement learning, instead of direct parsing 
of questions.
Both work use predefined module operations with handcrafted module  architectures. 
More recently, \cite{Johnson2017} proposes a model for visual reasoning that consists of a program generator and an execution engine.
The program generator constructs an explicit representation of the reasoning process to be performed. It is a sequence-to-sequence model which inputs the question as a sequence of words and outputs a program as a sequence of functions. 
The execution engine executes the resulting program to produce an answer. It is implemented using a neural module network. In contrast to~\cite{Andreas2016a,Andreas2016}, the modules use a generic architecture.
Similar to VQA, multi-domain image generation can also be regarded as a composition of several two domain image translations, which forms the bases of 
this paper.

\vspace{-0.15in}
\subsection{Image Translation}
\vspace{-0.05in}
\textbf{Generative Adversarial Networks} (GANs)~\cite{Goodfellow2014} are powerful generative models which have achieved impressive results in many computer vision tasks such as image generation\cite{Odena2016,Huang2017}, image inpainting~\cite{Iizuka2017}, super resolution~\cite{Ledig2017} and image-to-image translation~\cite{Isola2016,Li2016,Perarnau2016,Shen2017,Zhu2017}.
GANs formulate generative modeling as a game between two competing networks: a generator network produces synthetic data given some input noise and a discriminator network distinguishes between the generator's output and true data. 
The game between the generator $G$ and the discriminator $D$ has the minmax objective.
Unlike GANs which learn a mapping from a random noise vector to an output image, \textbf{conditional GANs} (cGANs)~\cite{Mirza2014} learn a mapping from a random noise vector to an output image conditioning on additional information. 
\textbf{Pix2pix}\cite{Isola2016} is a generic image-to-image translation algorithm using cGANs~\cite{Mirza2014}. It can produce reasonable results on a wide variety of problems. Given a training set which contains pairs of related images, pix2pix learns how to convert an image of one type into an image of another type, or vice versa.
\textbf{Cycle-consistent GANs} (CycleGANs)~\cite{Zhu2017} learn the image translation without paired examples. Instead, it trains two generative models cycle-wise between the input and output images. In addition to the adversarial losses, cycle consistency loss is used to prevent the two generative models from contradicting each other. 
Both Pix2Pix and CycleGANs are designed for two-domain image translation. 
By inverting the mapping of a cGAN~\cite{Mirza2014}, {\em i.e.}, mapping a real image into a latent space and a conditional representation, \textbf{IcGAN}~\cite{Perarnau2016} can reconstruct and modify an input image of a face conditioned on arbitrary attributes.
More recently, \textbf{StarGAN}~\cite{Choi2017} is proposed to perform multi-domain image translation using a single network conditioned on the target domain label. It learns the mappings among multiple domains using only a single generator and a discriminator.
Different from StarGAN, which learns all domain transformations within a single model, we train different simple composable translation networks for different attributes.

\vspace{-0.1in}
\section{Modular Generative Adversarial Networks}

\subsection{Problem Formulation}
We consider two types of multi-domain tasks: (i) {\em image generation} -- which directly generates an image with certain attribute properties from a random vector ({\em e.g.}, an image of a digit written in a certain font or style); and (ii) {\em image translation} -- which takes an existing image and minimally modifies it by changing certain attribute properties ({\em e.g.}, changing the hair color or facial expression in a portrait image).
%
%
We pre-define an attribute set $\mathbf{A}=\{A_1, A_2, \cdots, A_n\}$, where $n$ is the number of different attributes, and each attribute $A_i$ is a meaningful semantic property inherent in an image. For example, attributes for facial images may include hair color, gender or facial expression. Each $A_i$ has different attribute value(s), {\em e.g.}, 
black/blond/brown for hair color or male/female for gender.

For the {\em image generation} task, the goal is to learn a mapping $(z, \mathbf{a}) \mapsto y$. The input is a pair $(z, \mathbf{a})$, where $z$ is a randomly sampled vector and $\mathbf{a}$ is a subset of attributes $\mathbf{A}$. 
Note that the number of elements in $\mathbf{a}$ is not fixed; more elements would provide finer control over generated image.
The output $y$ is the target image.
For the {\em image translation} task, the goal is to learn a mapping $(x, \mathbf{a}) \mapsto y$. The input is a pair $(x, \mathbf{a})$, where $x$ is an image and $\mathbf{a}$ are the target attributes to be present in the output image $y$. 
The number of elements in $\mathbf{a}$ indicates the number of attributes of the input image that need to be altered. 

In the remainder of the section, we formulate the set of modules used for these two tasks and describe the process of composing them into networks.

\subsection{Network Construction}


\subsubsection{Image Translation.}
We first introduce the ModularGAN that performs multi-domain image translation. Four types of modules are used in this task: the encoder module ($\mathbf{E}$), which encodes an input image to an intermediate feature map; the transformer module ($\mathbf{T}$), which modifies a certain attribute of the feature map; the reconstructor module ($\mathbf{R}$), which reconstructs the image from an intermediate feature map; and the discriminator module ($\mathbf{D}$), which determines whether an image is real or fake, and predicts the attributes of the input image. More details about the modules will be given in the following section.

\begin{figure}[!t]
  \centering
  \includegraphics[width=1\columnwidth]{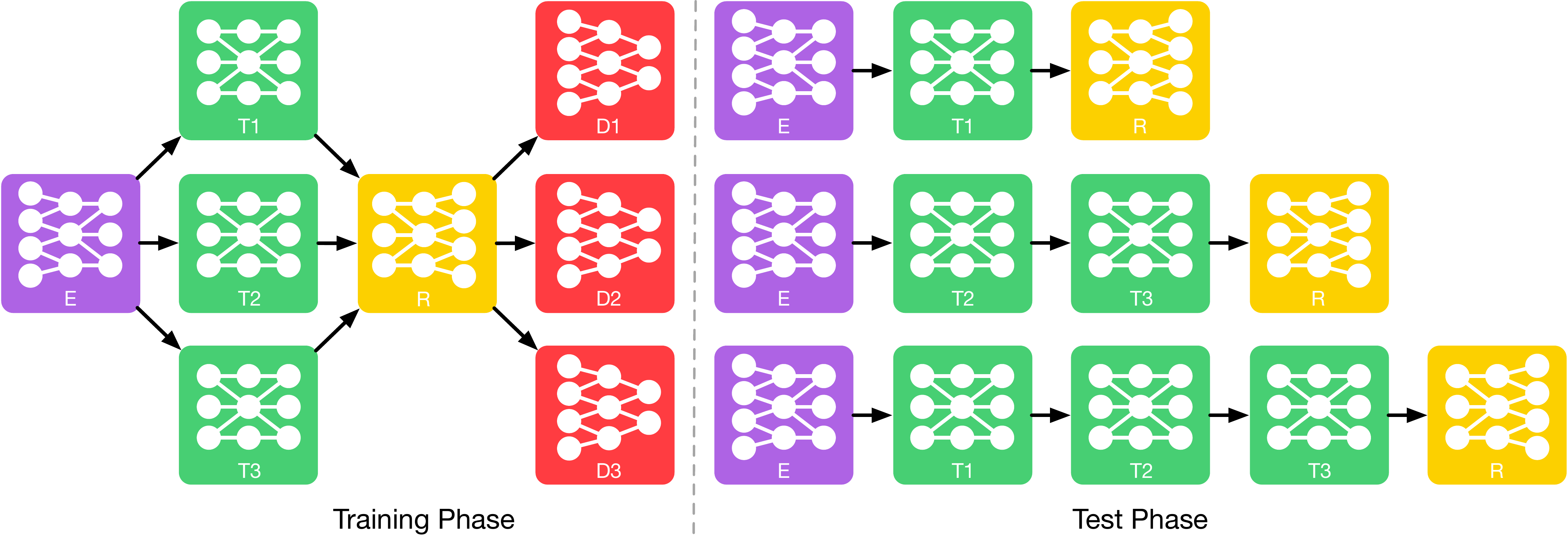}
  \caption{{\bf ModularGAN Architecture:} Multi-domain
  image translation architecture
  in training (left) and test (right) phases. ModularGAN consists of four different kinds of modules: the encoder module $\mathbf{E}$, transformer module $\mathbf{T}$, reconstructor module $\mathbf{R}$ and discriminator $\mathbf{D}$. These modules can be trained simultaneously and used to construct different generation networks according to the generation task in the test phase.}
  \vspace{-0.15in}
  \label{fig:arch}
\end{figure}

Fig.~\ref{fig:arch} demonstrates the overall architecture of the image translation model in the training and test phases. 
In the training phase (Fig.~\ref{fig:arch}, left), the encoder module $\mathbf{E}$ is connected to multiple transformer modules $\mathbf{T}_i$, each of which is further connected to a reconstructor module $\mathbf{R}$ to generate the translated image. There are multiple discriminator modules $\mathbf{D}_i$ connected to the reconstructor to distinguish the generated images from real images, and to make predictions of corresponding attribute. All modules have the same 
interface, {\em i.e.}, the output of $\mathbf{E}$, the input of $\mathbf{R}$, and both the input and output of $\mathbf{T}_i$ have the same shape and dimensionality. This enables the modules to be assembled in order to build more complex architectures at test time, as illustrated in Fig.~\ref{fig:arch}, right.  


In the training phase, an input image $x$ is first encoded by $\mathbf{E}$, which gives the intermediate representation $\mathbf{E}(x)$. Then different transformer modules $\mathbf{T}_i$ are applied to modify $\mathbf{E}(x)$ according to the pre-specified attributes $a_i$, resulting in $\mathbf{T}_i(\mathbf{E}(x), a_i)$. 
$\mathbf{T}_i$ is designed to transform a specific attribute $\mathbf{A}_i$ into a different attribute value\footnote{This also means that, in general, the number of transformer modules is equal to the number of attributes.}, {\em e.g.}, changing the hair color from blond to brown, or changing the gender from female to male. The reconstructor module $\mathbf{R}$ reconstructs the transformed feature map into an output image $y=\mathbf{R}(\mathbf{T}_i(\mathbf{E}(x), a_i))$. The discriminator module $\mathbf{D}$ is designed to distinguish the generated image $y$ and the real image $x$. It also predicts the attributes of the image $x$ or $y$. 

In the test phase (Fig.~\ref{fig:arch}, right), different transformer modules can be dynamically combined to form a network that can sequentially manipulate any number of attributes in arbitrary order. 



\vspace{-0.2in}
\subsubsection{Image Generation.}
The model architecture for the image generation task is mostly the same to the image translation task.
The only difference is that the encoder module $\mathbf{E}$ is replaced with a generator module $\mathbf{G}$, which generates an intermediate feature map $\mathbf{G}(z,a_0)$ from a random noise $z$ and a condition vector $a_0$ representing auxiliary information.
The condition vector $a_0$ could determine the overall content of the image. For example, if the goal is to generate an image of a digit, $a_0$ could be used to control which digit to generate, say digit 7.
A module $\mathbf{R}$ can similarly reconstruct an initial image $x = \mathbf{R}(\mathbf{G}(z,a_0))$, which is an image of digit 7 with any attributes. 
The remaining parts of the architecture are identical to the image translation task, which transform the initial image $x$ using a sequence of transformer modules $\mathbf{T}_i$ to alter certain attributes, ({\em e.g.},  color of the digit, stroke type or background).



\subsection{Modules}

\subsubsection{Generator Module ($\mathbf{G}$)} generates a feature map of size $C\times H \times W$ using several transposed convolutional layers. Its input is the concatenation of a random variable $z$ and a condition vector $a_0$. See supplementary materials for the network architecture.

\vspace{-0.2in}
\subsubsection{Encoder Module ($\mathbf{E}$)} encodes an input image $x$ into an intermediate feature representation of size $C \times H \times W$ using several convolutional layers. See supplementary materials for the network architecture.

\vspace{-0.2in}
\subsubsection{Transformer Module ($\mathbf{T}$)} 
\begin{figure}[!hbt]
  \centering
  \includegraphics[width=0.7\columnwidth]{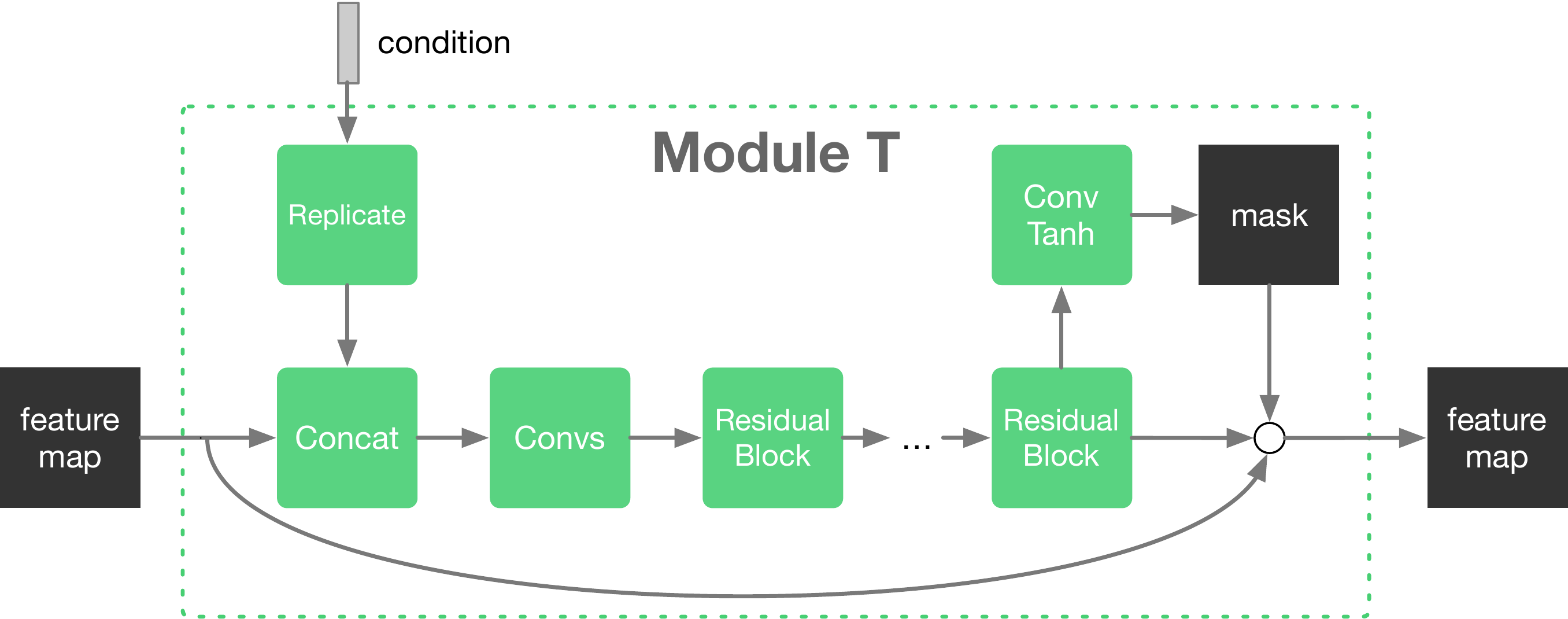}
  \caption{Transformer Module}
  \vspace{-0.2in}
  \label{fig:nett}
\end{figure}

is the core module in our model. It transforms the input feature representation into a new one according to input condition $a_i$.
A transformer module receives a feature map $f$ of size $C\times H \times W$ and a condition vector $a_i$ of length $c_i$. Its output is a feature map $f_t$ of size $C\times H \times W$.
Fig.~\ref{fig:nett} illustrates the structure of a module $\mathbf{T}$.
The condition vector $a_i$ of length $c_i$ is replicated to a tensor of size $c_i\times H\times W$, which is then concatenated with the input feature map $f$. 
Convolutional layers are first used to reduce the number of channels from $C + c_i$ to $C$. 
Afterwards, several residual blocks are sequentially applied, the output of which is denoted by $f'$.
Using the transformed feature map $f'$, additional convolution layers with the $Tanh$ activation function are used to generate a single-channel feature map $g$ of size $H\times W$. 
This feature map $g$ is further rescaled to the range $(0, 1)$ by $g' = (1+g)/2$.
The predicted $g'$ acts like an alpha mask or an attention layer: it encourages the module $\mathbf{T}$ to transform only the regions of the feature map that are relevant to the specific attribute transformation.
Finally, the transformed feature map $f'$ and the input feature map $f$ are combined using the mask $g'$ to get the output $f_t = g'\times f' + (1-g')\times f$.

\vspace{-0.2in}

\vspace{-0.2in}
\subsubsection{Reconstructor Module ($\mathbf{R}$)} reconstructs the image from a $C\times H \times W$ feature map using several transposed convolutional layers. See supplementary materials for the network architecture.


\vspace{-0.2in}
\subsubsection{Discriminator Module ($\mathbf{D}$)} classifies an image as real or fake, and predicts one of the attributes of the image ({\em e.g.}, hair color, gender or facial image). See supplementary materials for the network architecture.


\vspace{-0.1in}
\subsection{Loss Function}
We adopt a combination of several loss functions to train our model. 


\vspace{-0.2in}
\subsubsection{Adversarial Loss.}
We apply the adversarial loss~\cite{Goodfellow2014} to make the generated images look realistic. For the $i$-th transformer module $\mathbf{T}_i$ and its corresponding discriminator module $\mathbf{D}_i$, the adversarial loss can be written as:
\begin{align}
  \label{eq:adv_loss}
  \mathcal{L}_{\mathrm{adv}_i} 
  (\mathbf{E}, \mathbf{T}_i, \mathbf{R}, \mathbf{D}_i)
  = \ &\mathbb{E}_{y\sim p_{\mathrm{data}}(y)}[\log \mathbf{D}_i(y)] + \\\nonumber
  &\mathbb{E}_{x\sim p_{\mathrm{data}}(x)}[\log (1-\mathbf{D}_i(\mathbf{R}(\mathbf{T}_i(\mathbf{E}(x))))],
\end{align}
where $\mathbf{E}$, $\mathbf{T}_i$, $\mathbf{R}$, $\mathbf{D}_i$ are the encoder module, the $i$-th transformer module, the reconstructor module and the $i$-th discriminator module respectively.
$\mathbf{D}_{i}$ aims to distinguish between transformed samples $\mathbf{R}(\mathbf{T}_i(\mathbf{E}(x)))$ and real samples $y$. All the modules $\mathbf{E}$, $\mathbf{T}_i$ and $\mathbf{R}$ try to minimize this objective against an adversary $\mathbf{D}_i$ that tries to maximize it, \textit{i.e.} $\min_{\mathbf{E},\mathbf{T}_i,\mathbf{R}} \max_{\mathbf{D}_i}\mathcal{L}_{\mathrm{adv}_i}(\mathbf{E}, \mathbf{T}_i, \mathbf{R}, \mathbf{D}_i)$. 


\vspace{-0.2in}
\subsubsection{Auxiliary Classification Loss.}
Similar to \cite{Odena2016} and \cite{Choi2017}, for each discriminator module $\mathbf{D}_i$, besides a classifier to distinguish the real and fake images, we define an auxiliary classifier to predict the $i$-th attribute of the image, {\em e.g.},  hair color or  gender of the facial image. 
There are two components of the classification loss: real image loss $\mathcal{L}_{\mathrm{cls}_i}^r$ and fake image loss $\mathcal{L}_{\mathrm{cls}_i}^f$.

For real images $x$, the real image auxiliary classification loss $\mathcal{L}_{\mathrm{cls}_i}^r$ is defined as follows:
\begin{align}
  \label{eq:real_class}
  \mathcal{L}_{\mathrm{cls}_i}^r&=\mathbb{E}_{x,c_i}[-\log \mathbf{D}_{\mathrm{cls}_i}(c_i|x)],
\end{align}
where $\mathbf{D}_{\mathrm{cls}_i}(c|x)$ is the probability distribution over different attribute values predicted by $\mathbf{D}_i$, {\em e.g.}, black, blond or brown for hair color. 
The discriminator module $\mathbf{D}_i$ tries to minimize $\mathcal{L}_{\mathrm{cls}_i}^r$.

The fake image auxiliary classification loss $\mathcal{L}_{\mathrm{cls}_i}^f$ is defined similarly, using generated images $\mathbf{R}(\mathbf{E}(\mathbf{T}_i(x)))$:
\begin{align}
  \label{eq:fake_class}
  \mathcal{L}_{\mathrm{cls}_i}^f&=\mathbb{E}_{x,c_i}[-\log \mathbf{D}_{\mathrm{cls}_i}(c_i|\mathbf{R}(\mathbf{E}(\mathbf{T}_i(x))))].
\end{align}
The modules $\mathbf{R}$, $\mathbf{E}$ and $\mathbf{T}_i$ try to minimize $\mathcal{L}_{\mathrm{cls}_i}^f$ to generate fake images that can be classified as the correct target attribute $c_i$. 

\vspace{-0.2in}
\subsubsection{Cyclic Loss.}
Conceptually, the encoder module $\mathbf{E}$ and the reconstructor module $\mathbf{R}$ are a pair of inverse operations. Therefore, for a real image $x$, $\mathbf{R}(\mathbf{E}(x))$ should resembles $x$. Based on this observation, the encoder-reconstructor cyclic loss $\mathcal{L}_{\mathrm{cyc}}^{\mathbf{ER}}$ is defined as follows:
\begin{align}
    \label{eq:rec_E}
    \mathcal{L}_{\mathrm{cyc}}^{\mathbf{ER}}=\mathbb{E}_{x}[\|\mathbf{R}(\mathbf{E}(x))-x\|_1].
\end{align}

Cyclic losses can be defined not only on images, but also on intermediate feature maps.
At training time, different transformer modules $\mathbf{T}_i$ are connected to the encoder module $\mathbf{E}$ in a parallel fashion.
However, at test time $\mathbf{T}_i$ will be connected to each other sequentially, according to specific module composition for the test task.
Therefore it is important to have the cyclic consistency of the feature maps so that a sequence of $\mathbf{T}_i$ modifies the feature map consistently.
To enforce this, we define a cyclic loss on the transformed feature map and the encoded feature map of reconstructed output image. This cycle loss is defined as 
\begin{align}
    \label{eq:rec_Ti}
    \mathcal{L}_{\mathrm{cyc}}^{\mathbf{T}_i}=\mathbb{E}_{x}[\|\mathbf{T}_i(\mathbf{E}(x))-\mathbf{E}(\mathbf{R}(\mathbf{T}_i(\mathbf{E}(x))))\|_1],
\end{align}
where $\mathbf{E}(x)$ is the original feature map of the input image $x$,
and $\mathbf{T}_i(\mathbf{E}(x))$ is the transformed feature map.
The module $\mathbf{R}(\cdot)$ reconstructs the transformed feature map to a new image with the target attribute. The module $\mathbf{E}$ then encodes the generated image back to an intermediate feature map. This cyclic loss encourages the transformer module to output a feature map similar to the one produced by the encoder module. This allows different modules $\mathbf{T}_i$ to be concatenated at test time without loss in performance.

\vspace{-0.2in}
\subsubsection{Full Loss.}
Finally, the full loss functions for $\mathbf{D}$ is 
\begin{align}
    \mathcal{L}_D (\mathbf{D}) &= -\sum_{i=1}^{n}\mathcal{L}_{\mathrm{adv}_i} + \lambda_{\mathrm{cls}}\sum_{i=1}^{n}\mathcal{L}_{\mathrm{cls}_i}^r, \label{eq:loss_d}
\end{align}
and the full loss functions for 
$\mathbf{E}$, $\mathbf{T}$, $\mathbf{R}$ is 
\begin{align}
    \mathcal{L}_G (\mathbf{E}, \mathbf{T}, \mathbf{R}) &= \sum_{i=1}^{n}\mathcal{L}_{\mathrm{adv}_i} + \lambda_{\mathrm{cls}}\sum_{i=1}^{n}\mathcal{L}_{\mathrm{cls}_i}^f + \lambda_{\mathrm{cyc}}(\mathcal{L}_{\mathrm{cyc}}^{\mathbf{ER}} + \sum_{i=1}^{n}\mathcal{L}_{\mathrm{cyc}}^{\mathbf{T}_i}),
    \label{eq:loss_g}
\end{align}
where $n$ is the total number of controllable attributes, and $\lambda_{\mathrm{cls}}$ and $\lambda_{\mathrm{cyc}}$ are hyper-parameters that control the importance of auxiliary classification and cyclic losses, respectively, relative to the adversarial loss. 

\vspace{-0.1in}
\section{Implementation}
\vspace{-0.1in}
\subsubsection{Network Architecture.}
In our ModularGAN, $\mathbf{E}$ has two convolution layers with stride size of two for down-sampling. 
$\mathbf{G}$ has four transposed convolution layers with stride size of two for up-sampling.
$\mathbf{T}$ has two convolution layers with stride size of one and six residual block to transform the input feature map. Another convolution layer with stride size of one is added on top of the last residual block to predict a mask. 
$\mathbf{R}$ has two transposed convolution layers with stride size of two for up-sampling.
Five convolution layers with stride size of two are used in $\mathbf{D}$, together with two additional convolution layers to classify an image as real or fake, and its attributes.

\vspace{-0.2in}
\subsubsection{Training details.}
To stabilize the training process and to generate images of high quality, we replace the adversarial loss in Eq.~\eqref{eq:adv_loss} with the Wasserstein GAN~\cite{Arjovsky2017} objective function using gradient penalty~\cite{Gulrajani2017} defined by
\begin{align}
  \label{eq:wadv_loss}
  \mathcal{L}_{\mathrm{adv}_i} (\mathbf{E}, \mathbf{T}_i, \mathbf{R}, \mathbf{D}_i)
  =\ &\mathbb{E}_{x}[\mathbf{D}_i(x)] -\mathbb{E}_{x}[\mathbf{D}_i(\mathbf{R}(\mathbf{T}_i(\mathbf{E}(x))))]\\\nonumber
  &-\lambda_{\mathrm{gp}}\mathbb{E}_{\hat{x}}[(\|\triangledown_{\hat{x}} \mathbf{D}_i(\hat{x})\|_2-1)^2],
\end{align}
where $\hat{x}$ is sampled uniformly along a straight line between a pair of real and generated images. For all experiments, we set  $\lambda_{\mathrm{gp}}=10$ in Equation~\ref{eq:wadv_loss},  $\lambda_{\mathrm{cls}}=1$ and $\lambda_{\mathrm{cyc}}=10$ in Equation~\ref{eq:loss_d} and Equation~\ref{eq:loss_g}. We use the Adam optimizer~\cite{Kingma2013} with a batch size of 16. All networks are trained from scratch with an initial learning rate of 0.0001. We keep the same learning rate for the first 10 epochs and linearly decay the learning rate to 0 over the next 10 epochs.
\vspace{-0.1in}
\section{Experiments}
\vspace{-0.1in}
We first conduct image generation experiments on a synthesized multi-attribute MNIST dataset. Next, we compare our method with recent work on image-to-image facial attributes transfer. Our method shows both qualitative and quantitative improvements as measured by 
user studies and attribute classification. 
Finally, we conduct an  ablation study to examine the effect of mask prediction in module $\mathbf{T}$, the cyclic loss, and the order of multiple modules $\mathbf{T}$ on multi-domain image transfer.

\vspace{-0.2in}
\subsection{Baselines}
\vspace{-0.05in}
\noindent
{\bf IcGAN} first learns a mapping from a latent vector $z$ to a real image $y$, $G: (z,c) \mapsto y$, then learns the inverse mapping from a real image $x$ to a latent vector $z$ and a condition representation $c$, $E: x \mapsto (z,c)$. Finally, it reconstructs a new image conditioned on $z$ and a modified $c'$, i.e. $G: (z,c') \mapsto {y}$.

\vspace{0.05in}
\noindent
{\bf CycleGAN} learns two mappings $G: x \mapsto y$ and $F: y \mapsto x$ simultaneously, and uses a cycle consistency loss to enforce $F(G(x))\approx x$ and $G(F(y))\approx y$. 
We train different models of CycleGAN for each pair of domains in our experiments.

\vspace{0.05in}
\noindent
{\bf StarGAN} trains a single $G$ to translate an input image $x$ into an output image $y$ conditioned on the target domain label(s) $c$ directly, {\em i.e.}, $G: (x, c) \mapsto y$. Setting multiple entries in $c$ allows StarGAN to perform multi-attribute transfer.

\vspace{-0.1in}
\subsection{Datasets}


\subsubsection{ColorMNIST.} 
We construct a synthetic dataset called the ColorMNIST, based on the MNIST Dialog Dataset~\cite{Seo2017}. Each image in ColorMNIST contains a  digit with four randomly sampled attributes, \textit{i.e.}, $\mathrm{number} = \{x \in \mathbb{Z}|0 \leqslant x \leqslant 9\}$, $\mathrm{color} = \{red, blue, green, purple, brown\}$, $\mathrm{style} = \{flat,stroke\}$, and $\mathrm{bgcolor} =\{cyan, yellow, white,silver,salmon\}$. 
We generate 50K images of size $64\times64$.
 
\vspace{-0.15in}
\subsubsection{CelebA.} 
The CelebA dataset~\cite{Liu2015} contains 202,599 face images of celebrities, with 40 binary attributes such as young, smiling, pale skin and male. 
We randomly sampled 2,000 images as test set and use all remaining images as training data. All images are center cropped with size 178 $\times$ 178, and  resized to 128$\times$128. We choose three attributes with seven different attribute values for all the experiments: hair color = $\{black, blond, brown\}$, gender = $\{male, female\}$, and smile = $\{smile, no\-smile\}$.


\vspace{-0.15in}
\subsection{Evaluation}

\subsubsection{Classification Error.}
As a quantitative evaluation, we compute the classification error of each attribute on the synthesized images using a ResNet-18 network~\cite{He2016}, which is trained to classify the attributes of an image.
All methods use the same classification network for performance evaluation. 
Lower classification errors imply that the generated images have more accurate target attributes.



\vspace{-0.2in}
\subsubsection{User Study.}
We also perform a user study using Amazon Mechanical Turk (AMT) to assess the image quality for image translation tasks. 
Given an input image, the Turkers were instructed to choose the best generated image based on perceptual realism, quality of transfer in attribute(s), and preservation of a figure’s original identity. 

\vspace{-0.15in}
\subsection{Experimental Results on ColorMNIST}

\begin{figure}[!t]
\begin{center}
\includegraphics[width=1\columnwidth]{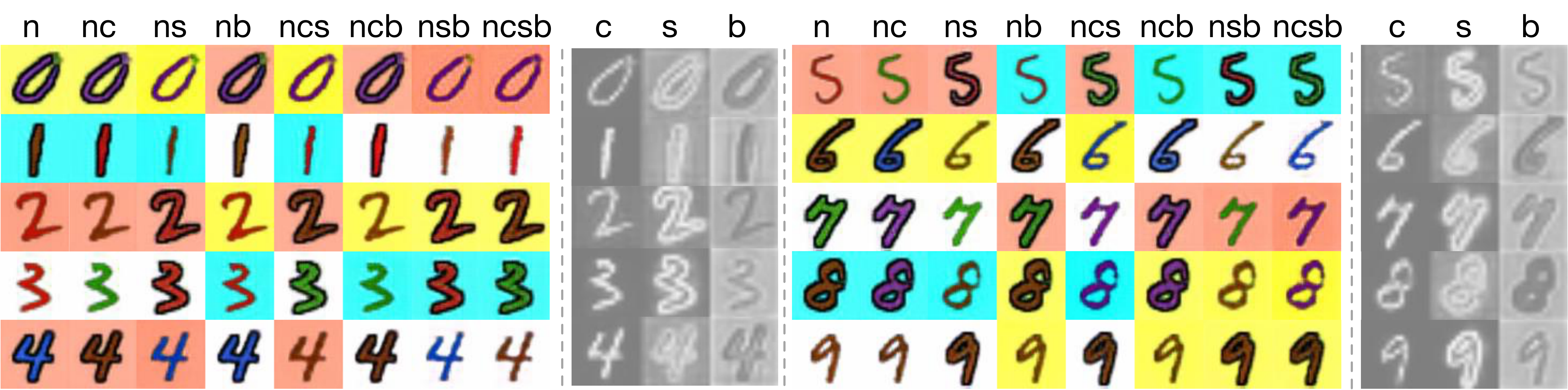}
\end{center}
\vspace{-0.2in}
\caption{{\bf Image Generation:} Digits synthesis results on the ColorMNIST dataset. Note, that (n) implies conditioning on the digit number, (c) color, (s) stroke type, and (b) background. Columns denoted by more than one letter illustrate generation results conditioned on multiple attributes, {\em e.g.}, (ncs) -- digit number, color, and stroke type. Greayscale images illustrate mask produced internally by $\mathbf{T}_i$ modules, $i \in \{c,s,b\}$.}
\vspace{-0.2in}
\label{fig:results_colormnist}
\end{figure}

\subsubsection{Qualitative evaluation.}
Fig.~\ref{fig:results_colormnist} shows the digit image generation results on ColorMNIST dataset.
The generator module $\mathbf{G}$ and reconstructor module $\mathbf{R}$ first generate the correct digit according to the number attribute as shown in the first column. The generated digit has random color, stroke style and background color.
By passing the feature representation produced by $\mathbf{G}$ through different $\mathbf{T}_i$, 
the digit color, stroke style and background of the initially generated image will change,  as shown in the second to forth columns.
The last four columns illustrate multi-attribute transformation by combining different $\mathbf{T}_i$. 
Each module $\mathbf{T}_i$ only changes a specific attribute and keeps other attributes untouched (at the previous attribute value). Note that there are scenarios where the initial image already has the target attribute value; in such cases the transformed image is identical to the previous one.

\vspace{-0.15in}
\subsubsection{Visualization of Masks.}
In Fig.~\ref{fig:results_colormnist}, we also visualize the predicted masks in  each transformer module $\mathbf{T}_i$. It provides an interpretable way to understand where the modules apply the transformations. 
White pixels in the mask correspond to regions in the feature map that are modified by the current module;
black pixels to regions that remain unchanged throughout the module.
It can be observed that the color transformer module $\mathbf{T}_c$ mainly 
changes the interior of the digits,
so only the digits are highlighted. The stroke style transformer module $\mathbf{T}_s$ correctly focuses on the borders of the digits. Finally, the masks corresponding to the background color transformer module $\mathbf{T}_b$ have larger values in the background regions.

\vspace{-0.1in}
\subsection{Experimental Results on CelebA}

\subsubsection{Qualitative evaluation.}

\begin{figure}[!t]
\begin{center}
\includegraphics[width=0.95\columnwidth]{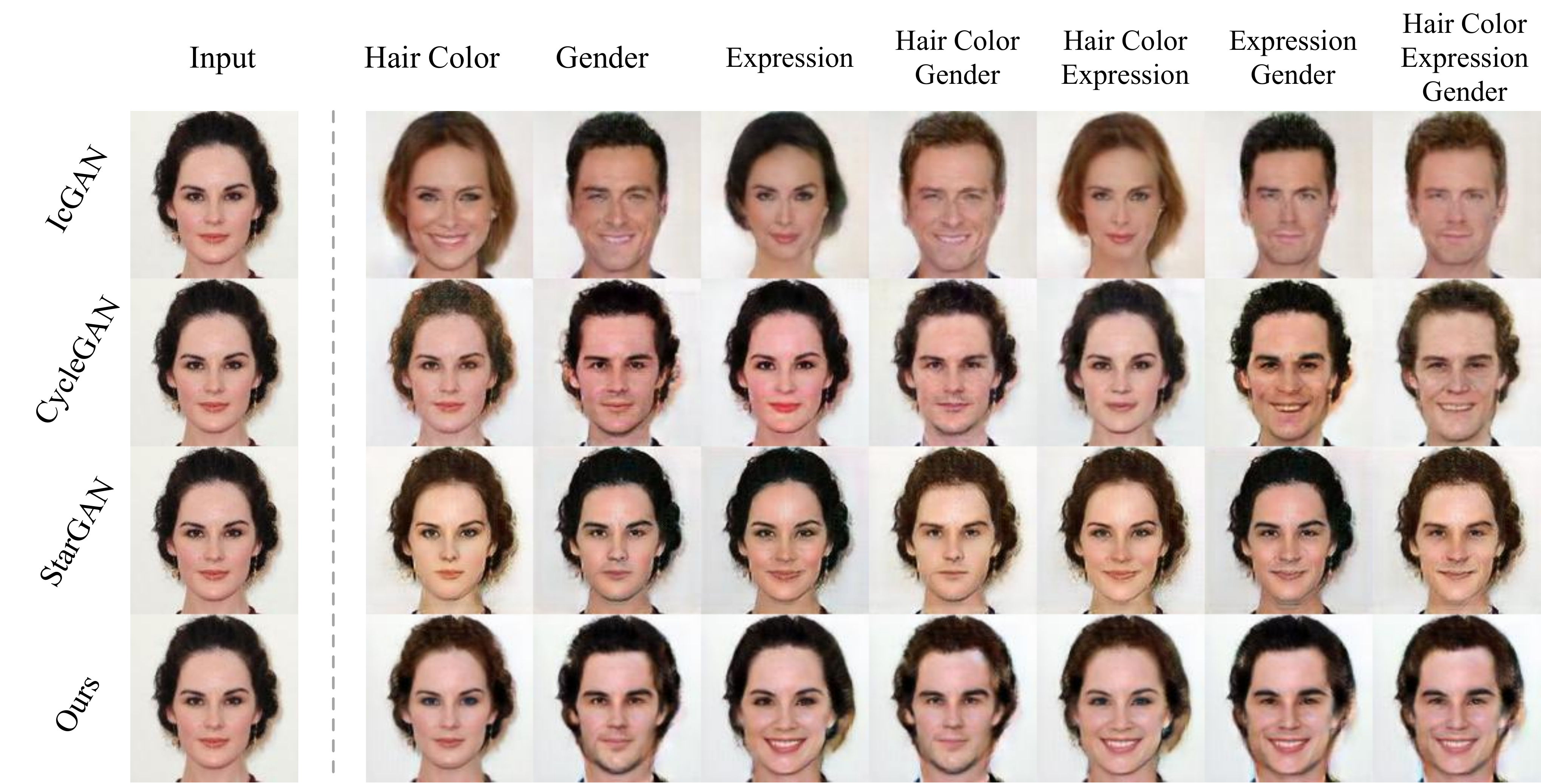}
\end{center}
   \vspace{-0.2in}
   \caption{{\bf Facial attribute transfer results on CelebA:} See text for description.}
  \vspace{-0.1in}
   \label{fig:results_celeba}
\end{figure}

Fig.~\ref{fig:results_ours} and Fig.~\ref{fig:results_celeba} show the facial attribute transfer results on CelebA  using the proposed method and the baseline methods, respectively. In Fig.~\ref{fig:results_celeba}, the transfer is between a female face image with neutral expression and black hair to a variety of combinations of attributes. 
The results show that IcGAN has the least satisfying performance. Although the generated images have the desired attributes, the facial identity is not well preserved. The generated images also do not have sharp details, caused by the information lost during the process of encoding the input image into a low-dimensional latent vector and decoding it back. 
The images generated by CycleGAN are better than IcGAN, but there are some visible artifacts. By using the cycle consistence loss, CycleGAN preserves the facial identity of the input image and only changes specific regions of the face.
StarGAN generates better results than CycleGAN, since it is trained on the whole dataset and implicitly leverages images from all attribute domains.
Our method generates better results than the baseline methods ({\em e.g.}, see Smile or multi-attribute transfer in the last column). 
It uses multiple transformer modules to change different attributes, and each transformer module learns a specific mapping from one domain to another. This is different from StarGAN, which learns all the transformations in one single model.

\vspace{-0.15in}
\subsubsection{Visualization of Masks.}
\begin{figure}[!t]
\begin{center}
\includegraphics[width=0.85\columnwidth]{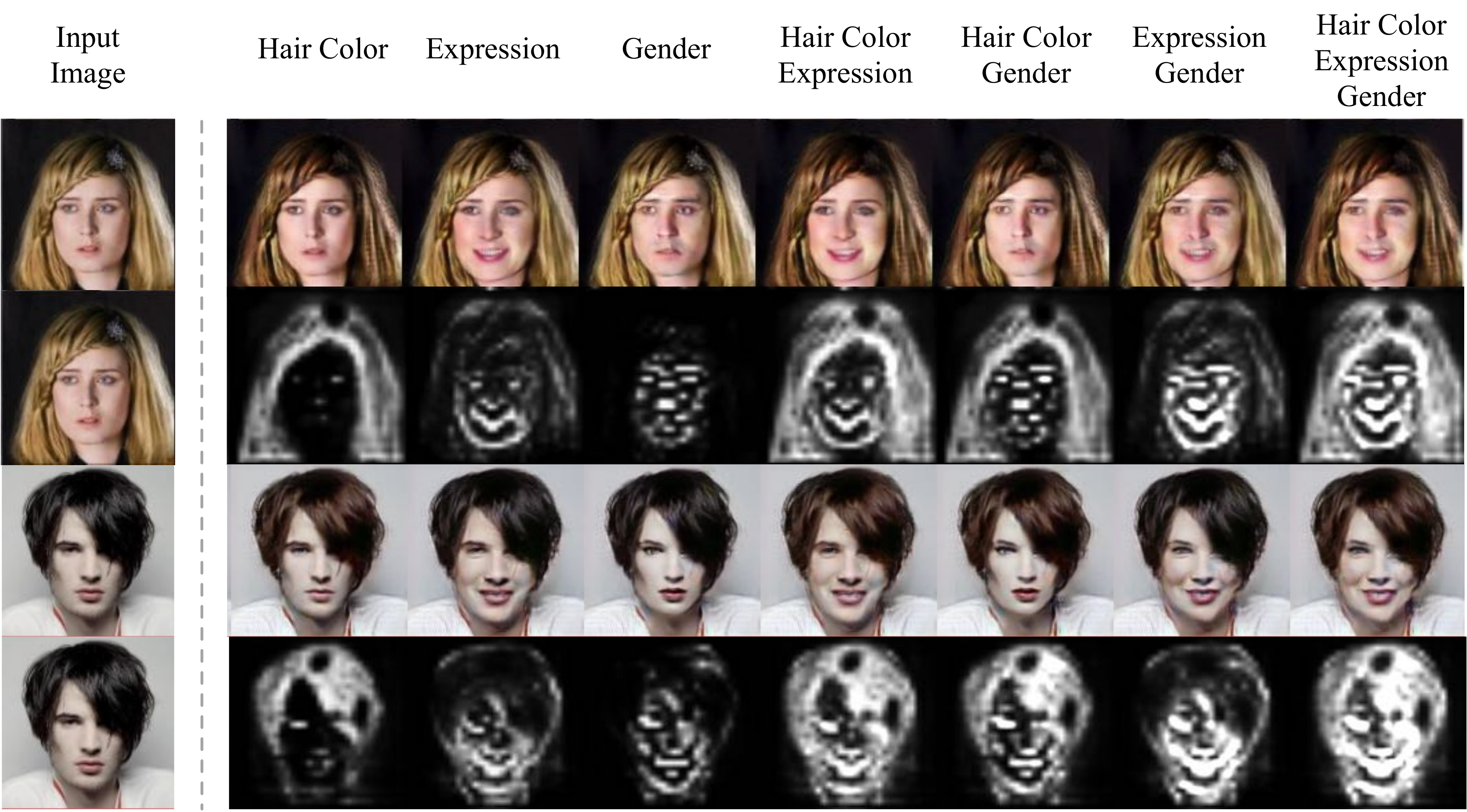}
\end{center}
   \vspace{-0.15in}
   \caption{{\bf Mask Visualization:} Visualization of masks when performing attribute transfer. We sum the different masks when multiple modules $\mathbf{T}$ are used.}
   \label{fig:results_mask_celeba}
\end{figure}

To better understand what happens when ModularGAN translates an image, we visualize the mask of each transformer module in Fig.~\ref{fig:results_mask_celeba}. When multiple $\mathbf{T}_i$ are used, we add different predicted masks. 
It can be seen from the visualization that when changing the hair color, the transformer module only focuses on the hair region of the image. 
By modifying the mouth area of the feature maps, the facial expression can be changed from neutral to smile. 
To change the gender, regions around cheeks, chin and nose are used.

\vspace{-0.15in}
\subsubsection{Quantitative evaluation.}
\vspace{-0.25in}
\begin{table}[!tbh]
\footnotesize
\centering
\caption{{\bf AMT User Study:} Higher values are better and indicating preference.}
\label{tb:is_baselines}
\begin{tabular}{|c|c|c|c|c|c|c|c|}
\hline
Method   & H     & S     & G     & HS    & HG    & SG    & HSG   \\ \hline
IcGAN    & 3.48  & 2.63  & 8.70  & 4.35  & 8.70  & 13.91 & 15.65 \\ \hline
CycleGAN  & 17.39 & 16.67 & 29.57 & 18.26 & 20.00 & 17.39 & 9.57  \\ \hline
StarGAN  & 30.43 & 36.84 & \textbf{32.17} & 31.30 & 27.83 & 27.83 & 27.83 \\ \hline
\textbf{Ours}     & \textbf{48.70} & \textbf{43.86} & 29.57 & \textbf{46.09} & \textbf{43.48} & \textbf{40.87} & \textbf{46.96} \\ \hline
\end{tabular}
\end{table}

\vspace{-0.4in}
\begin{table}[!bbh]
\footnotesize
\centering
\caption{{\bf Classification Error:} Lower is better, indicating fewer attribute errors.}
\label{tb:clss_baselines}
\begin{tabular}{|c|c|c|c|c|c|c|c|}
\hline
Method        & H             & S             & G             & HS            & HG            & SG            & HSG           \\ \hline
IcGAN         & 7.82          & 10.43         & 20.86         & 22.17         & 20.00         & 23.91         & 23.18         \\ \hline
CycleGAN      & 4.34          & 10.43         & 13.26         & 13.67         & 10.43         & 17.82         & 21.01         \\ \hline
StarGAN       & \textbf{3.47} & 4.56          & 4.21          & 4.65          & 6.95          & 5.52          & 7.63          \\ \hline
\textbf{Ours} & 3.86          & \textbf{4.21} & \textbf{2.61} & \textbf{4.03} & \textbf{6.51} & \textbf{4.04} & \textbf{6.09} \\ \hline
\end{tabular}
\end{table}
\vspace{-0.2in}

We train a model that classifies the hair color, facial expression and gender on the CelebA dataset using a ResNet-18 architecture~\cite{He2016}. The training/test set are the same as that in other experiments.
The trained model classifies the hair color, gender and smile with accuracy of 96.5\%, 97.9\% and 98.3\% respectively. 
We then apply this trained model on transformed images produced by different methods
on the test set. 
As can be seen in Table~\ref{tb:clss_baselines}, our model achieves a comparable classification error to StarGAN on the hair color task, and the lowest classification errors on all other tasks. This indicates that our model produces realistic facial images with desired attributes.
Table~\ref{tb:is_baselines} shows the results of the AMT experiments. Our model obtains the majority of votes for best transferring attributes in all the cases except gender. 
We observe that our gender transfer model better preserves original hair, which is desirable from the model's point of view, but sometimes perceived negatively by the Turkers. 

\subsection{Ablation Study}

To analyze the effect of the mask prediction, the cyclic loss and the order of modules $\mathbf{T}_i$ when transferring multiple attributes, we conduct ablation experiments by removing the mask prediction, removing the cyclic loss and randomizing the order of $\mathbf{T}_i$. 

\begin{figure}[!tbh]
\begin{center}
\includegraphics[width=0.95\columnwidth]{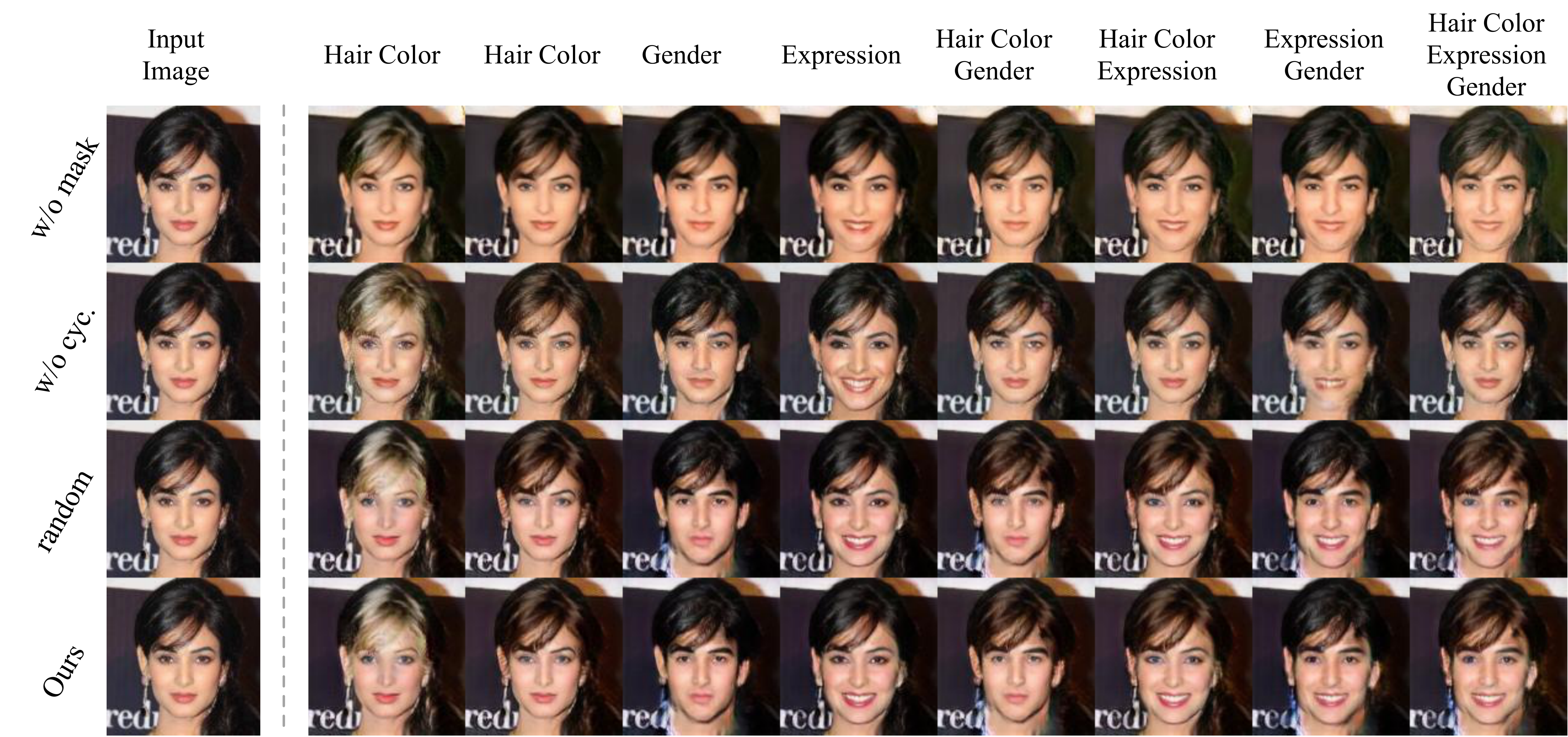}
\end{center}
\vspace{-0.2in}
   \caption{{\bf Ablation:} Images generated using different variants of our method. From top to bottom: ModularGAN w/o mask prediction in $\mathbf{T}$, ModularGAN w/o cyclic loss, ModularGAN with random order of $\mathbf{T}_i$ when performing multi-attribute transfer.}
\label{fig:results_ablation}
\end{figure}

\vspace{-0.15in}
\subsubsection{Effect of Mask.}

Fig.~\ref{fig:results_ablation} shows that, without mask prediction, the model can still manipulate the images but tends to perform worse on gender, smile and multi-attribute transfer. 
Without the mask, $\mathbf{T}$ module not only needs to learn how to translate the feature map, but also needs to learn how to keep parts of the original feature map intact.
As a result, without mask it becomes difficult to compose modules, as illustrated by higher classification errors reported in Table~\ref{tb:class_ab}. 

\vspace{-0.15in}
\subsubsection{Effect of Cyclic Loss.}
Removing the cyclic loss does not affect the results of single-attribute manipulation, as shown in Fig.~\ref{fig:results_ablation}. However, when combining multiple transformer modules, the model can no loner generate
images with desired attributes. 
This is also quantitatively verified in Table~\ref{tb:class_ab}: the performance of multi-attribute transfer drops dramatically without the cyclic loss.

\vspace{-0.15in}
\subsubsection{Effect of Module Order}
We test our model by applying $\mathbf{T}_i$ modules in random order when performing multi-attribute transformations (as compared to fixed ordering - Ours). The results reported in Table~\ref{tb:class_ab} indicate that our model is unaffected by the order of transformer modules, which is a desired property. 


\begin{table}[!tbh]
\footnotesize
\centering
\vspace{-0.1in}
\caption{{\bf Ablation Results:} Classification error for ModularGAN variants (see text).}
\vspace{-0.1in}
\label{tb:class_ab}
\begin{tabular}{|l|c|c|c|c|c|c|c|}
\hline
\multicolumn{1}{|c|}{Method} & H             & S             & G             & HS            & HG            & SG            & HSG           \\ \hline
Ours w/o mask                & 4.01          & 4.65          & 3.58          & 30.85         & 34.67         & 36.61         & 56.08         \\ \hline
Ours w/o cyclic loss                 & 3.93          & 4.48          & 2.87          & 25.34         & 28.82         & 30.96         & 52.87         \\ \hline
Ours random order            & \textbf{3.86} & \textbf{4.21} & \textbf{2.61} & 4.37          & \textbf{5.98} & 4.13          & 6.23          \\ \hline
\textbf{Ours}                & \textbf{3.86} & \textbf{4.21} & \textbf{2.61} & \textbf{4.03} & 6.51          & \textbf{4.04} & \textbf{6.09} \\ \hline
\end{tabular}
\vspace{-0.2in}
\end{table}

\vspace{-0.1in}
\section{Conclusion}
\vspace{-0.1in}
In this paper, we proposed a novel modular multi-domain generative adversarial network architecture, which consists of several reusable and composable modules. Different modules can be jointly trained end-to-end efficiently.
By utilizing the mask prediction within module $\mathbf{T}$ and the cyclic loss, different (transformer) modules can be combined in order to successfully translate the image to different domains.
Currently, different modules are connected sequentially in test phase. Exploring different structure of modules for more complicated tasks will be one of our future work directions. 

\section*{Acknowledgement}
We gratefully acknowledge the support of NVIDIA Corporation with the donation of the Titan Xp GPU used for this research.

\bibliographystyle{splncs}
\bibliography{egbib}

\clearpage

\chapter*{Supplementary Materials}
\vspace{-0.4in}

\section*{Network Architecture}
The network architectures of different modules used in ModularGAN are shown in Table~\ref{tb:nete}, Table~\ref{tb:netg}, Table~\ref{tb:nett}, Table~\ref{tb:netr} and Table~\ref{tb:netd}, respectively. There are some notations: $h$ and $w$ are the height and width of input image, $c_i$: the number of different value of $i$-th attribute, N: the number of output channels, K: kernel size, S: stride size, P: padding size, IN: instance normalization. 

\begin{table}[]
\scriptsize
\centering
\caption{Architecture of Encoder Module $\mathbf{E}$}
\label{tb:nete}
\begin{tabular}{c|c|c}
\hline
Part         & Input $\to$ Output Shape                                              & Layer Information                          \\ \hline\hline
\multirow{3}{*}{Down-sampling}& $(h, w, 3) \to (h, w, 64)$                                            & CONV-(N64, K7, S1, P3), IN, ReLU  \\ 
& $(h, w, 64) \to (\frac{h}{2}, \frac{w}{2}, 128)$                      & CONV-(N128, K4, S2, P1), IN, ReLU \\ 
& $(\frac{h}{2}, \frac{w}{2}, 128) \to (\frac{h}{4}, \frac{w}{4}, 256)$ & CONV-(N256, K4, S2, P1), IN, ReLU \\ \hline
\multirow{6}{*}{Residual Blocks} & $(\frac{h}{4}, \frac{w}{4}, 256) \to (\frac{h}{4}, \frac{w}{4}, 256)$     & CONV-(N256, K3, S1, P1), IN, ReLU                    \\
& $(\frac{h}{4}, \frac{w}{4}, 256) \to (\frac{h}{4}, \frac{w}{4}, 256)$     & CONV-(N256, K3, S1, P1), IN, ReLU \\
& $(\frac{h}{4}, \frac{w}{4}, 256) \to (\frac{h}{4}, \frac{w}{4}, 256)$     & CONV-(N256, K3, S1, P1), IN, ReLU \\
& $(\frac{h}{4}, \frac{w}{4}, 256) \to (\frac{h}{4}, \frac{w}{4}, 256)$     & CONV-(N256, K3, S1, P1), IN, ReLU \\
& $(\frac{h}{4}, \frac{w}{4}, 256) \to (\frac{h}{4}, \frac{w}{4}, 256)$     & CONV-(N256, K3, S1, P1), IN, ReLU \\
& $(\frac{h}{4}, \frac{w}{4}, 256) \to (\frac{h}{4}, \frac{w}{4}, 256)$     & CONV-(N256, K3, S1, P1), IN, ReLU \\\hline
\end{tabular}
\end{table}

\begin{table}[]
\scriptsize
\centering
\vspace{-0.2in}
\caption{Architecture of Generator Module $\mathbf{G}$}
\label{tb:netg}
\begin{tabular}{c|c|c}
\hline
Part         & Input $\to$ Output Shape                                              & Layer Information                          \\ \hline\hline
\multirow{4}{*}{Up-sampling}& $(1, 1, z+c_0) \to (\frac{h}{32}, \frac{w}{32}, 2048)$                                            & DECONV-(N2048, K4, S1, P0), IN, ReLU  \\
& $(\frac{h}{32}, \frac{w}{32}, 2048) \to (\frac{h}{16}, \frac{w}{16}, 1024)$                      & DECONV-(N1024, K4, S2, P1), IN, ReLU \\ 
& $(\frac{h}{16}, \frac{w}{16}, 1024) \to (\frac{h}{8}, \frac{w}{8}, 512)$                      & DECONV-(N512, K4, S2, P1), IN, ReLU \\
& $(\frac{h}{8}, \frac{w}{8}, 512) \to (\frac{h}{4}, \frac{w}{4}, 256)$ & DECONV-(N256, K4, S2, P1), IN, ReLU \\ \hline
\end{tabular}
\end{table}


\begin{table}[!h]
\scriptsize
\centering
\vspace{-0.2in}
\caption{Architecture of Transformer Module $\mathbf{T}$}
\label{tb:nett}
\begin{tabular}{c|c|c}
\hline
Part                             & Input $\to$ Output Shape                                                  & Layer Information                                          \\\hline\hline
Transform Layer                      & $(\frac{h}{4}, \frac{w}{4}, 256+c_i) \to (\frac{h}{4}, \frac{w}{4}, 256)$ & CONV-(N256, K3, S1, P1), IN, ReLU                 \\\hline
\multirow{6}{*}{Residual Blocks} & $(\frac{h}{4}, \frac{w}{4}, 256) \to (\frac{h}{4}, \frac{w}{4}, 256)$     & CONV-(N256, K3, S1, P1), IN, ReLU                    \\
& $(\frac{h}{4}, \frac{w}{4}, 256) \to (\frac{h}{4}, \frac{w}{4}, 256)$     & CONV-(N256, K3, S1, P1), IN, ReLU \\
& $(\frac{h}{4}, \frac{w}{4}, 256) \to (\frac{h}{4}, \frac{w}{4}, 256)$     & CONV-(N256, K3, S1, P1), IN, ReLU \\
& $(\frac{h}{4}, \frac{w}{4}, 256) \to (\frac{h}{4}, \frac{w}{4}, 256)$     & CONV-(N256, K3, S1, P1), IN, ReLU \\
& $(\frac{h}{4}, \frac{w}{4}, 256) \to (\frac{h}{4}, \frac{w}{4}, 256)$     & CONV-(N256, K3, S1, P1), IN, ReLU \\
& $(\frac{h}{4}, \frac{w}{4}, 256) \to (\frac{h}{4}, \frac{w}{4}, 256)$     & CONV-(N256, K3, S1, P1), IN, ReLU \\\hline
Mask Prediction Layer          & $(\frac{h}{4}, \frac{w}{4}, 256) \to (\frac{h}{4}, \frac{w}{4}, 1)$       & CONV-(N1, K7, S1, P3), TanH\\\hline                 
\end{tabular}
\end{table}

\begin{table}[!h]
\scriptsize
\centering
\vspace{-0.2in}
\caption{Architecture of Generator Module $\mathbf{R}$}
\label{tb:netr}
\begin{tabular}{c|c|c}
\hline
Part         & Input $\to$ Output Shape                                              & Layer Information                          \\ \hline\hline
\multirow{3}{*}{Up-sampling}& $(\frac{h}{4}, \frac{w}{4}, 256) \to (\frac{h}{2}, \frac{w}{2}, 128)$                                            & CONV-(N128, K7, S1, P3), IN, ReLU  \\
& $(\frac{h}{2}, \frac{w}{2}, 128) \to (h, w, 64)$                      & CONV-(N64, K4, S2, P1), IN, ReLU \\
& $(h, w, 64) \to (h, w, 3)$ & CONV-(N3, K4, S2, P1), IN, ReLU \\ \hline
\end{tabular}
\end{table}


\begin{table}[!h]
\scriptsize
\centering
\vspace{-0.1in}
\caption{Architecture of Discriminator Module $\mathbf{D}$}
\label{tb:netd}
\begin{tabular}{c|c|c}
\hline
Part         & Input $\to$ Output Shape                                              & Layer Information                          \\ \hline\hline
Input Layer  & $(h, w, 3) \to (\frac{h}{2}, \frac{w}{2}, 64)$                                     & CONV-(N64, K4, S2, P1), IN, Leaky ReLU  \\ \hline
\multirow{5}{*}{Hidden Layer} & $(\frac{h}{2}, \frac{w}{2}, 64) \to (\frac{h}{4}, \frac{w}{4}, 128)$                      & CONV-(N128, K4, S2, P1), IN, Leaky ReLU \\ 
& $(\frac{h}{4}, \frac{w}{4}, 128) \to (\frac{h}{8}, \frac{w}{8}, 256)$                      & CONV-(N256, K4, S2, P1), IN, Leaky ReLU \\
& $(\frac{h}{8}, \frac{w}{8}, 256) \to (\frac{h}{16}, \frac{w}{16}, 512)$                      & CONV-(N512, K4, S2, P1), IN, Leaky ReLU \\
& $(\frac{h}{16}, \frac{w}{16}, 512) \to (\frac{h}{32}, \frac{w}{32}, 1024)$                      & CONV-(N1024, K4, S2, P1), IN, Leaky ReLU \\
& $(\frac{h}{32}, \frac{w}{32}, 64) \to (\frac{h}{64}, \frac{w}{64}, 2048)$                      & CONV-(N2048, K4, S2, P1), IN, Leaky ReLU \\
\hline
Output Layer ($\mathbf{D}$) & $(\frac{h}{64}, \frac{w}{64}, 2048) \to (\frac{h}{64}, \frac{w}{64}, 1)$ & CONV-(N1, K4, S2, P1), IN, Leaky ReLU \\
Output Layer ($\mathbf{D}_{{cls}}$) & $(\frac{h}{64}, \frac{w}{64}, 2048) \to (1, 1, c_i)$ & CONV-(N($c_i$), K4, S2, P1), IN, Leaky ReLU \\\hline
\end{tabular}
\vspace{-0.2in}
\end{table}

\section*{Additional Qualitative Results on CelebA Dataset}
\vspace{-0.2in}
\begin{figure}[!b]
	\centering
	\vspace{-0.2in}
	\includegraphics[width=0.95\columnwidth]{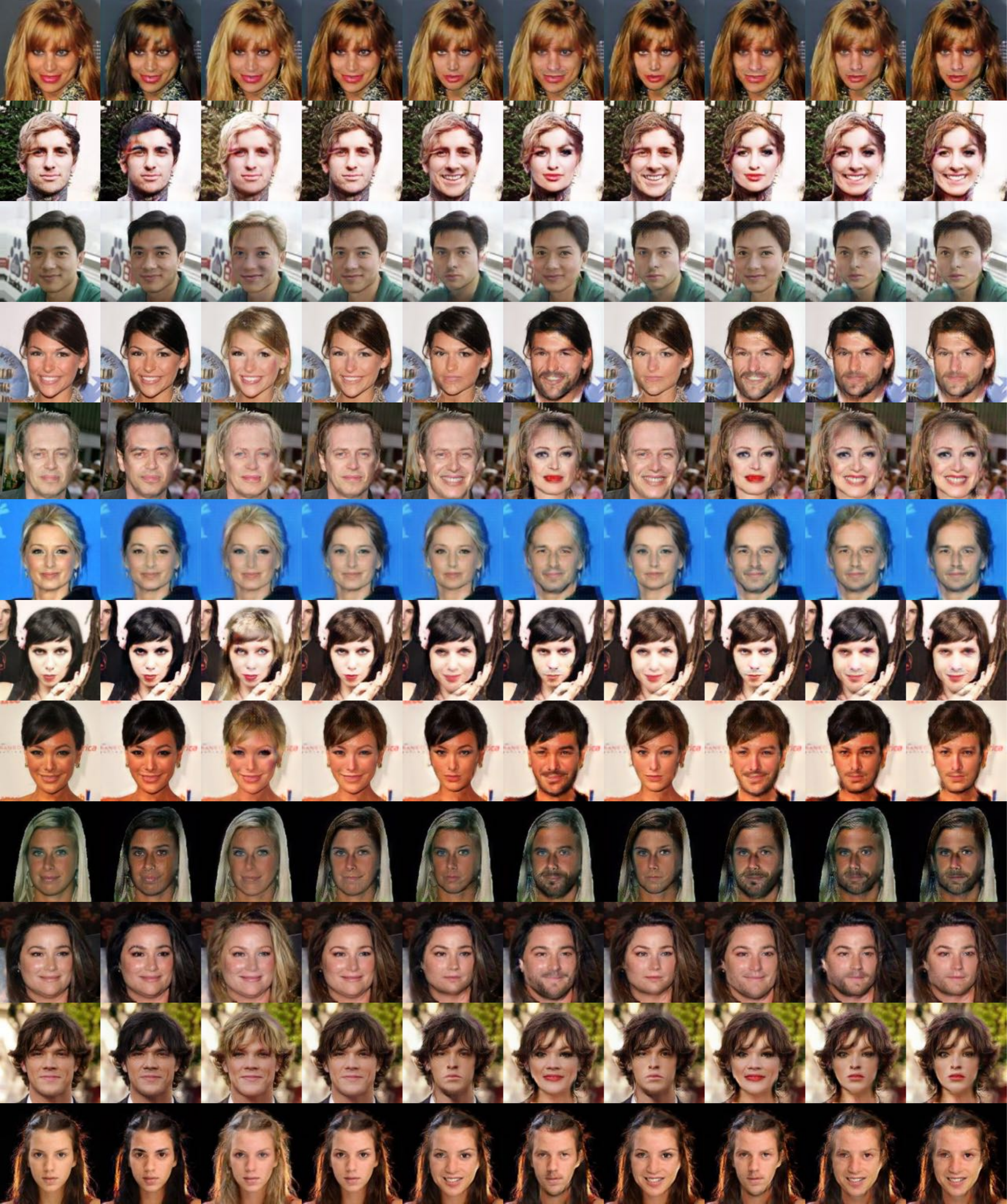}
	\caption{Single and multiple attribute transfer on CelebA (Input, Black hair, Blond hair, Brown hair, Expression, Gender, Hair color + Expression, Hair color + Gender, Expression + Gender, Hair color + Expression + Gender)}
	\vspace{-0.2in}
\end{figure}
\end{document}